\documentclass[letterpaper, 10 pt, journal, twoside]{IEEEtran}  % Comment this line out if you need a4paper

\IEEEoverridecommandlockouts                              % This command is only needed if 
\usepackage{graphicx} % for pdf, bitmapped graphics files
\usepackage[table, dvipsnames]{xcolor}
\usepackage{amsmath} % assumes amsmath package installed
\usepackage{amssymb}  % assumes amsmath package installed
\usepackage{acronym}
\usepackage{todonotes}
\usepackage[font=small]{caption}
\usepackage{subcaption}
\usepackage{array,booktabs}
\usepackage{url}
\usepackage{hyperref}
\usepackage[noabbrev]{cleveref}
\usepackage{pbox}
\usepackage{scalerel}
\usepackage{siunitx}

\graphicspath{{fig/}}

\newcommand{\x}{\mathbf{x}}
\newcommand{\y}{\mathbf{y}}
\newcommand{\z}{\mathbf{z}}
\renewcommand{\c}{\mathbf{c}}
\newcommand{\sun}{\textit{Sun}}
\newcommand{\rain}{\textit{Rain}}
\newcommand{\twilight}{\textit{Twilight}}
\newcommand{\wet}{\textit{Wet}}
\newcommand{\fire}{\textit{Fire}}
\newcommand{\tpr}{TPR @ 5\% FPR}

\newcommand{\edited}[1]{#1}

\acrodef{CNN}{convolutional network}
\acrodef{NN}{neural network}
\acrodef{AUROC}{area under the receiver-operator characteristic curve}
\definecolor{maxperf}{rgb}{0,0.66,0}
\definecolor{minperf}{rgb}{1,0.0,0}

\title{
Safe Robot Navigation via Multi-Modal Anomaly Detection
}

\author{Lorenz Wellhausen$^{1}$, Ren{\'e} Ranftl$^{2}$ and Marco Hutter$^{1}$% <-this % stops a space
\thanks{Manuscript received: September, 11th, 2019; Revised December, 6th, 2019; Accepted January, 2nd, 2020.}%Use only for final RAL version
\thanks{This paper was recommended for publication by Editor Eric Marchand upon evaluation of the Associate Editor and Reviewers' comments.}
\thanks{This work was supported by the Intel Network on Intelligent Systems, the Swiss National Science Foundation
(SNF) through the NCCR Robotics, and the EU Horizon 2020 research and innovation programme (No 780883). 
It has been conducted as part of ANYmal Research, a community to advance legged robotics}% <-this % stops a space
\thanks{$^{1}$Robotic Systems Lab, ETH Z{\"u}rich, Switzerland}%
\thanks{$^{2}$Intel Labs, Munich, Germany}%
\thanks{Digital Object Identifier (DOI): see top of this page.}
}

\begin{document}
\maketitle
% \thispagestyle{empty}
% \pagestyle{empty}

%===============================================================================

\begin{abstract}

Navigation in natural outdoor environments requires a robust and reliable traversability classification method to handle the plethora of situations a robot can encounter.
Binary classification algorithms perform well in their native domain but tend to provide overconfident predictions when presented with out-of-distribution samples, which can lead to catastrophic failure when navigating unknown environments. 
We propose to overcome this issue by using anomaly detection on multi-modal images for traversability classification, which is easily scalable by training in a self-supervised fashion from robot experience.
In this work, we evaluate multiple anomaly detection methods with a combination of uni- and multi-modal images in their performance on data from different environmental conditions. 
Our results show that an approach using a feature extractor and normalizing flow with an input of RGB, depth and surface normals performs best. It achieves over 95\% area under the ROC curve and is robust to out-of-distribution samples.
\end{abstract}


%===============================================================================

\section{Introduction}
\label{sec:introduction}

\IEEEPARstart{R}{obot} navigation through natural outdoor environments introduces challenges which are usually not considered when deploying autonomous systems in indoor and man-made environments. 
The most notable difference is that perceived geometry can not be assumed to be rigid. 
The implications for this are two-fold:
First, while flat terrain is typically assumed to be traversable, it can actually be untraversable or dangerous for robot navigation if the terrain is non-rigid. 
Treacherous terrain like deep sand, mud and bodies of water show flat geometry but are potentially fatal for many robots.
Second, while obstacles are often simply considered as the presence of geometry, this does not hold when a robot can "push through" a compliant obstruction. 
Vegetation like grass and small bushes are difficult to identify from purely geometric information but are frequently encountered in natural environments.

This implies that semantic environment information is desirable, if not necessary, in addition to geometric information to navigate such environments.
While analytical models have been successfully used to infer traversability from geometric information~\cite{wermelinger2016navigation}, deriving an analytical model for semantic information from image data is infeasible, due to the high dimensionality of the problem. 

Machine-learning models achieve state-of-the-art performance in semantic image processing, using manually labelled data~\cite{valada2017adapnet, romera2018erfnet}. 
However, manually labelling data is cumbersome and not scalable to larger quantities of data. 
Additionally, it relies on a human expert who often lacks a good intuition about traversability for environments where the robot has not been operated before and cannot provide quantitative terrain information. 

When collecting self-supervised samples through robot experience, which we have shown in previous work~\cite{wellhausen2019selfsupervised}, positive samples for traversable terrain can be gathered safely and in large quantities. 
Collecting negative samples, however, implies provoking robot failure which can be harmful for the robot.
In addition, labelling negative samples can never cover the entire domain of untraversable terrains and possible obstructions, which can lead to over-confident classifier output when presented with out-of-training-distribution samples.  

Detection of out-of-distribution samples, also called anomaly detection, or novelty detection, has received increased attention with the recent success of deep learning in general, and semantic image processing specifically.
These methods can be trained using positive samples only, and are by design robust to out-of-distribution samples.
However, existing work does not fully commit to the concept~\cite{kerner2019novelty,richter2017safe}, assumes constant appearance of the environment~\cite{christiansen2016deepanomaly}, and doesn't leverage geometric information.
In this work, we present an approach to fully leverage anomaly detection using appearance and geometric information for safe robot navigation in various environments. 

\begin{figure}[t]
  \includegraphics[width=\linewidth]{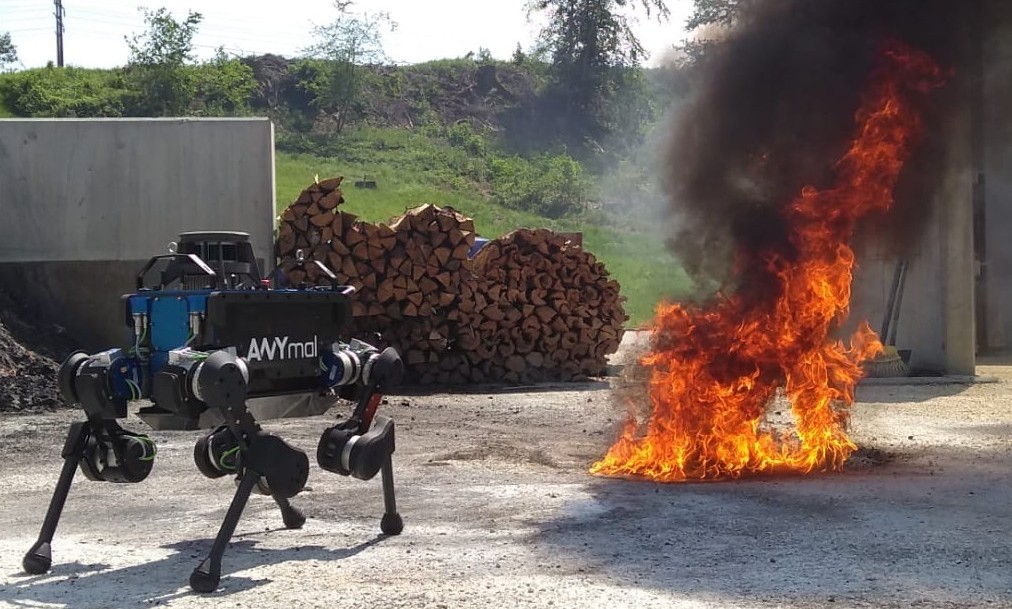}
  \caption{Anomaly detection allows robots to operate in environments with unforeseen and rarely occurring obstacles.}
  \label{fig:wangen_fire}
\end{figure}

The main contribution of this work is an extensive evaluation of multiple anomaly detection methods and sensor modality combinations with respect to their performance on self-supervised data.
We release the dataset used in this work to enable reproduction of  our results and to develop the concept of anomaly-based navigation further. 
Lastly, we combine ideas from other works~\cite{ruff2018deep, blum2019fishyscapes} into a new anomaly detection approach, which trains a feature embedding directly by maximizing the log-likelihood.

We show that we are able to train an anomaly detection method using only positive examples of multi-modal data to be highly discriminative. 
Our best model reaches more than 95\% \ac{AUROC}, which enables safe robot navigation.

%===============================================================================

\section{Related Work}
\label{sec:related_work}

Traditional navigation approaches for mobile robots use a geometric environment representation as their only basis for traversability estimation~\cite{wermelinger2016navigation, chavez2017image, krusi2017driving, oleynikova2017voxblox}.
This line of work is well developed and provides good performance in man-made environments, but fails to capture compliant terrain.

Semantic-aware navigation approaches typically leverage additional sensor modalities to infer additional terrain information~\cite{rothrock2016spoc, kim2006traversability, bradley2015scene, otsu2016autonomous, barnes2017find, hirose2018gonet, valada2017adapnet, cunningham2017improving, ordonez2018modeling}.
Approaches using more unconventional sensors either require a long observation duration~\cite{cunningham2017improving} or a bulky sensor payload~\cite{ordonez2018modeling} which exceeds the capabilities of our target platform.
Therefore, most work is focused on camera-based methods and either performs semantic segmentation of the environment~\cite{rothrock2016spoc,bradley2015scene,otsu2016autonomous,valada2017adapnet} or directly predicts a traversability label~\cite{kim2006traversability,barnes2017find,hirose2018gonet}.
Semantic segmentation approaches~\cite{bradley2015scene,valada2017adapnet} can perform well in environments similar to their training domain but do not transfer to unknown environments~\cite{blum2019fishyscapes}.
While this is not prohibitive in some domains~\cite{rothrock2016spoc,otsu2016autonomous}, in most cases even small changes in the environmental conditions, for example due to weather, can drastically change the appearance of terrain classes.

In previous work~\cite{wellhausen2019selfsupervised} we have shown that we can predict terrain properties ahead of the robot without classifying the terrain. 
This can be used to make informed navigation decisions on terrain that is traversable, but does not provide a traversability classification itself. 
We therefore require an additional method to provide traversability labels, which should also be learned in a self-supervised fashion to maintain scalability of the navigation pipeline.

Some recent work has proposed weakly- and self-supervised learning for navigation purposes by combining multiple sensors~\cite{otsu2016autonomous,barnes2017find} or proprioception~\cite{kim2006traversability,hirose2018gonet}.
All of these approaches use binary classification with a fixed set of pre-defined classes, however. 
This has shown to be prone to overconfident predictions in the presence of out-of-distribution data~\cite{hein2019relu}, which can lead to disastrous consequences~\cite{bozhinoski2019safety}. 

Anomaly detection could solve this problem by learning the distribution of safe terrain, which makes the approach more robust to out-of-distribution samples.
Numerous work is available for anomaly detection, which uses autoencoders~\cite{haselmann2018anomaly}, support vectors~\cite{scholkopf2000support, ruff2018deep}, generative adversarial networks\cite{schlegl2017unsupervised} and normalizing flow~\cite{choi2018generative}.

%Some other work has recently explored navigation via anomaly detection.
Anomaly detection has been used for indoor navigation~\cite{richter2017safe}, planetary exploration~\cite{kerner2019novelty}, and for navigation in agricultural fields~\cite{christiansen2016deepanomaly}.
However, these approaches are either reduced in scope~\cite{richter2017safe}, rely on a consistent terrain appearance~\cite{christiansen2016deepanomaly}, or use an additional binary classifier to make the final anomaly decision~\cite{kerner2019novelty}.
 
We propose a scalable approach for safe navigation which can be trained in a fully self-supervised fashion from only traversable examples.
We learn the distribution of terrain which the robot has safely traversed before and consider out-of-distribution samples as unsafe.
This enables safe robot locomotion, even in the presence of unknown obstacles.

%===============================================================================

\section{Method}
\label{sec:method}

We aim to learn a model of the typical appearance of terrain that \edited{the robot has safely navigated before.} 
We can use this model for safe navigation by classifying new sensory inputs into "known" and "unknown" terrain classes.
We use an automated pipeline that automatically generates positive labels from sensory data. 
We then evaluate the performance of different novelty detection methods and input modalities in various scenarios. 
We further briefly outline how the resulting image-based labels can be used in a robot navigation framework.

\subsection{Data Collection}\label{sec:data}

We collect positive terrain samples from robot-experience in a self-supervised fashion. 
The basic pipeline was presented in our earlier work~\cite{wellhausen2019selfsupervised}.
The quadrupedal robot ANYmal is teleoperated over various terrain while we record the image streams of an onboard camera, as well as the foothold contact locations in a robot-centered frame.
We use Visual SLAM on the image stream to recover the camera poses and foothold locations in a common coordinate frame.
This allows us to project all footholds along the robot trajectory into all camera images. 
We consequently obtain image locations which correspond to positive labels for traversability.
In a final step, we extract the image patches at the foothold locations to generate our training dataset.

Our pipeline can be applied to any dense exteroceptive sensor. 
in this work we use a RGB-D camera to sense both appearance and 
geometry of the environment.
% In the experiment section we will show that multi-modal data is highly benefitial for anomaly detection.
We hereafter refer to \textit{images} as the stack of RGB and depth images and potentially derived quantities.

\subsection{Anomaly Detection}\label{sec:anomaly_detection}

We evaluate multiple approaches with respect to to their anomaly detection performance on our data.
Our investigation is focused on deep learning approaches with fully convolutional architectures, since they achieve state-of-the-art performance, and are efficient during test time even on larger input images.

We define a feature encoder ${f(\x)\rightarrow \y}$, which maps an image patch $\x\in\mathbb{R}^{w\times h\times c}$ of width $w$ and height $h$ and a channel depth of $c$ to a feature vector $\y\in\mathbb{R}^d$. 
The encoder architecture will be shared by all 
novelty detection approaches.
%We call the function $f(\x)$ encoder, or feature generator.
%Because we use small image patches at train time, the output is one feature vector per image patch.
We implement $f(\x)$ with a fully-convolutional neural network in order to support inference on arbitrary image sizes ($W\times H$). 
We will obtain an output tensor $\y_{\text{inf}} \in \mathbb{R}^{j \times k \times d}$ with $j\approx\frac{W}{4}$ and $k\approx\frac{H}{4}$, which we use for localized anomaly detection in the full-size image.

The network architecture for the encoder uses three consecutive blocks, each consisting of a convolutional layer with kernel size 5 with a leaky ReLU non-linearity.
The first two blocks are followed by a MaxPool layer of size 2, while the last block is followed by a final convolution with kernel size 1.
The number of channels is, in sequence, $[c, 32, 64, 128, 128]$, where $c$ is the number of input channels.

We further denote the training loss as $\mathcal{L}$ and the anomaly decision criterion as $\mathcal{C}$. 
\edited{We use a simple threshold} on the decision criterion to classify patches into their respective classes.

\subsubsection{Autoencoder~\cite{sakurada2014anomaly}}
Autoencoders are neural networks that consist of an encoder $f(\x)$ to generate a (low-dimensional) latent feature vector $\y$ from the image patch and a decoder $f'(\y)$, which tries to reconstruct the input patch from this latent vector.
Since the feature vector is low-dimensional when compared to the dimensionality of the input patch, an internal information bottleneck is introduced. 
The autoencoder is thus forced to learn descriptive image features in order to be able to reconstruct the input.
In the context of anomaly detection, the basic assumption is that the autoencoder will over-fit to the training distribution. 
Anomalous input images will therefore be reconstructed with less accuracy than images that are similar to the training images. 

Our implementation uses a decoder network $f'(\y)$ that is composed of convolution layers of the same dimensions as the encoder, but with nearest-neighbor upscaling layers instead of MaxPooling.
The training loss for the autoencoder is given by
\begin{equation}
    \mathcal{L}_{\text{AE}}(\y) = \mathcal{C}_{\text{AE}}(\y) = \frac{1}{n} \sum_n (f'(\y) - \x)^2.
\end{equation}
Note, that we use the reconstruction error in image space as the decision criterion.

\subsubsection{Deep SVDD}
Ruff et al. \cite{ruff2018deep} propose an anomaly
detection approach based on deep networks.
In this approach, a neural network is trained to extract image features that are contained in a hypersphere, where the hypersphere
is jointly adapted during training of the feature extractor. 
At test time, samples which fall outside of the hypersphere are assumed to be anomal samples. 
Ruff et al. \cite{ruff2018deep} propose two different variants of this general idea. A soft-boundary formulation with the training loss
\begin{equation}
    \mathcal{L}_{\text{Soft}}(\y) = R^2 + \frac{1}{\nu} \max\{0, \lVert \y - \c\rVert_2^2 - R^2\},
\end{equation}
and a hard-boundary formulation with loss
\begin{equation}
    \mathcal{L}_{\text{Hard}}(\y) = \lVert \mathbf{y - \c} \rVert_2^2.
\end{equation}

The center of the hypersphere $\c \in \mathbb{R}^d$ is an arbitrary, fixed, non-zero vector that needs to be chosen in advance. 
We follow the recommendations of the original authors and initialize it with an initial forward-pass on the untrained network~\cite{ruff2018deep}.
Note, that the decision radius $R$ in the soft-boundary formulation is optimized together with the parameters of the feature generator.

We use the squared distance to the center of the hypersphere as decision criterion for both formulations:
\begin{equation}
    \mathcal{C}_{\text{SVDD}} = \lVert \y - \c \rVert_2^2.
\end{equation}

\subsubsection{Embedding + Real-NVP}

\begin{figure}[t]
  \includegraphics[width=\linewidth]{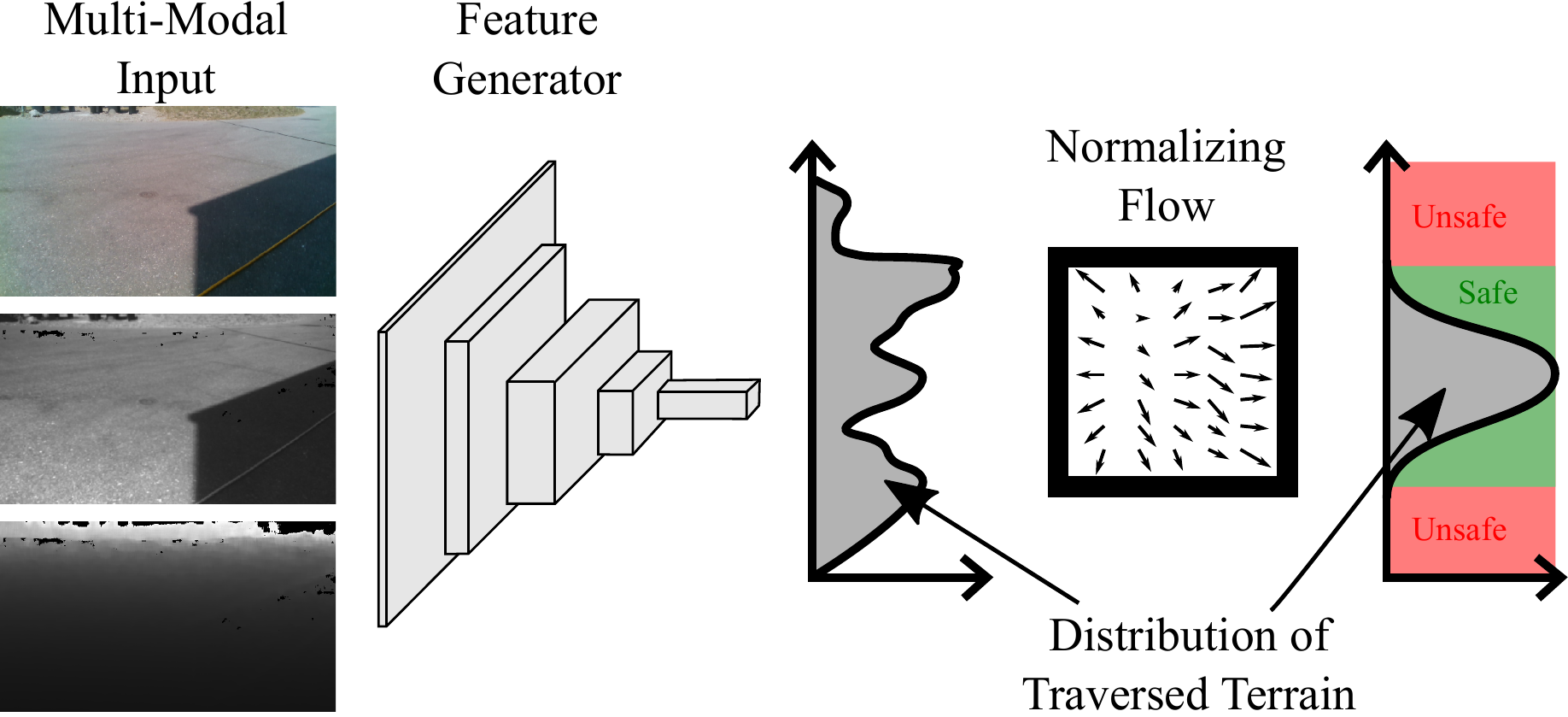}
  \caption{Multi-modal input images are projected into a feature space to form a distribution of safe terrain features. Normalizing flow is used to transform this distribution and facilitate exact likelihood computation.}
  \label{fig:approach}
\end{figure}

Normalizing flow models~\cite{rezende2015variational,dinh2016density,kingma2018glow} are powerful methods which can be used to learn arbitrary probability distributions by maximizing the likelihood of training samples. 
The normalizing flow approach is inherently probabilistic. 
As a consequence, it naturally handles variations and noise in the input data. 
Furthermore, since the likelihood is a metric for how likely it is that a given distribution has generated a given feature, it is a very natural decision criterion for anomaly detection.
 
Since normalizing flow methods limit themselves to be composed of specific, invertible network modules, they allow for tractable computation of the log-determinant of their Jacobian.
However, this restriction combined with a high-dimensional feature space when operating on images makes deep architectures and large amounts of training data necessary. 
This makes them impractical to use for our application. 

Similar to a method proposed by Blum et al.~\cite{blum2019fishyscapes}, we combine normalizing flows with a convolutional embedding network, which generates a lower-dimensional feature vector from an image patch. 
We then learn the safe terrain distribution in the low-dimensional latent space. 
A schematic overview of the approach is depicted in \Cref{fig:approach}.

We use  Real-NVP~\cite{dinh2016density} as our normalizing flow method.
Let $g(\y)\rightarrow \z$ be a bijection, which transforms the latent variable $\y$ into another vector $\z\in\mathbb{R}^d$ of same dimensionality.
We assume a given prior distribution $p_Z(\z)$ on the transformed vector $\z$. 
The prior distribution can be chosen arbitrarily, as long as its log-likelihood can easily be computed.
Using the change-of-variable formula we can obtain the log-likelihood of the posterior distribution in latent space, which serves as our loss function:
\begin{equation}
    \mathcal{L}_{\text{NVP}}(\y) = -\log(p_Z(g(\y))) - \log(|\det\Big(\frac{\delta g(\y)}{\delta \y^T}\Big)|).
\end{equation}

Real-NVP specifically limits the modules to scaling and translation of the intermediate features. 
The log-determinant can consequently be computed as the sum of the scaling factors~\cite{dinh2016density}.
We further directly use the log-likelyhood as the decision criterion:

\begin{equation}
 \mathcal{C}_{\text{NVP}}(\y) = \mathcal{L}_{\text{NVP}}(\y).
\end{equation}

Our Real-NVP flow network has 6 affine coupling layers where scaling and translation coefficients are obtained from a MLP with two hidden layers.

\subsection{Input Modalities}

Intel RealSense cameras provide RGB, infrared and depth streams. 
We ignore the infrared stream in this work, due to inconsistencies between infrared imagers of different camera models.
However, in addition to RGB and raw depth, we also consider modalities that are derived from depth and the robot state information: 

\subsubsection{Gravity Aligned Depth}
We project the depth image into 3D space using the camera intrinsics $\mathbf{K}$ and then rotate these points with the orientation of the camera in the gravity-aligned odometry frame $\mathbf{R}_{oc} \in SO(3)$ which is provided by the inertial-kinematic state estimator. 
We then combine the two horizontal axes into the distance in the horizontal plane. %under the assumption that only the absolute distance from the robot is informative for anomaly detection, but not the direction.
Let $d$ be the depth value at image coordinates $[u,v]$.
\begin{equation}
\begin{split}
    \mathbf{p} & = \mathbf{R}_{oc}\cdot \mathbf{K}^{-1}\cdot
    \begin{bmatrix}
        u \\ v \\ 1
    \end{bmatrix} \cdot d, \\
    \mathbf{d}_g & = 
    \begin{bmatrix}
        d_{\text{horz}} \\
        d_{\text{vert}}
    \end{bmatrix}
    =
    \begin{bmatrix}
        \sqrt{p_x^2 + p_y^2} \\
        p_z
    \end{bmatrix}.
\end{split}
\label{eq:depth_3d}
\end{equation}

\subsubsection{Gravity Aligned Surface Normals}
We compute gravity-aligned surface normals $\mathbf{n}_g$ from $\mathbf{p}$ using the FALS algorithm~\cite{badino2011fast} and combine horizontal components in the same way we did for $\mathbf{d}_g$:
\begin{equation}
    \mathbf{n}_g = 
    \begin{bmatrix}
        n_{\text{horz}} \\
        n_{\text{vert}}
    \end{bmatrix}
    =
    \begin{bmatrix}
        \sqrt{n_x^2 + n_y^2} \\
        n_z
    \end{bmatrix}.
\end{equation}

\subsubsection{Surface Normal Angle}
We compute the angle between surface normal and the horizontal plane $n_{\text{ang}}$ as 
\begin{equation}
    n_{\text{ang}} = \arctan \Big( \frac{n_{\text{vert}}}{n_{\text{horz}}} \Big).
\end{equation}

\subsection{Navigation}

All anomaly detection methods are trained on image patches and are fully convolutional.
This means we can deploy them on larger images than they were trained on to obtain an anomaly mask for the input image. 
We use this mask as a measure of traversability.
We then use the depth channel of our input image to project the anomaly mask to 3D space, which gives us point estimates for traversability in 3D.
Finally, these measurements can be used in a mapping framework to obtain a environment representation that can be used for planning. 
In our case, we opted for a 2D grid representation, which is common for ground robots and can be used for efficient path planning.
% This approach is schematically shown in \Cref{fig:navigation}.

%===============================================================================

\section{Experimental Results}
\label{sec:result}

Experiments were performed on data collected with the ANYmal~\cite{hutter2016anymal} quadruped, with image data captured using Intel Realsense cameras. 
Data for the base training set was captured on a Realsense ZR300, while test data was recorded on a Realsense D435. 
ANYmal was teleoperated over various terrain, with the forward-facing cameras at a slight downward angle, which varied between sorties.

We provide code and dataset online to reproduce our results and to encourage further research on anomaly navigation.\footnote{\url{http://github.com/leggedrobotics/anomaly_navigation}}

\subsection{Dataset}

\edited{We use the data collected for our previous work~\cite{wellhausen2019selfsupervised} as our base training set, which represents about $2.5$ hours of continuous robot operation under sunny and overcast lighting conditions. 
It was collected by teleoperating the robot through an urban park, a forest, and farmland, and covers various terrain types like asphalt, grass, dirt and sand.}

We also recorded new data in a search-and-rescue training facility to evaluate this work.
\edited{We chose this particular training site, because we can artificially create anomalous obstacles and events in a safe and controlled fashion.
Note that this method is not specific to search-and-rescue scenarios and can be used for general-purpose navigation.}
In this new location, the robot followed the same general path multiple times under different environmental conditions.

\begin{enumerate}
    \item \sun: Direct sunlight in the afternoon.
    \item \fire: Direct sunlight in the afternoon, but with a controlled fire in the robot field-of-view.
    \item \rain: During rain, with varying intensity from light to moderately heavy rain.
    \item \wet: In the late afternoon under direct sunlight, with wet ground from preceding rain.
    \item \twilight: Just after sunset during twilight. 
\end{enumerate}

For \sun{} and \twilight{} we each performed two sorties following the same path, while we could only perform a shorter second sortie for \rain{} and no second sortie for \wet{} and \fire{}.

As network input at training time, we choose image patches of size $32\times32$.
Patches of traversable terrain are extracted in a self-supervised fashion, as described in \Cref{sec:data}.
While image patches of traversable terrain are sufficient to train our approaches, patches of untraversable terrain are necessary for a quantitative performance analysis.
Because we do not have self-supervised data of untraversable terrain we manually label 500 negative samples in each sortie for our evaluation.
Note that we do not need to manually label any training data, as all approaches are trained using positive samples only. 
Through this approach we obtained $10\,000$ training image patches and around $500$ positive and negative samples for each test sortie.

\subsection{Network Training}
Both Deep-SVDD (\textit{SVDD Soft} + \textit{SVDD Hard}) and Real-NVP (\textit{NVP}) methods are evaluated with randomly initialized weights (\textit{No Pretraining}), as well as with the feature generator pretrained using the autoencoder (\textit{Pretrained}). 
For the Real-NVP architecture we also tried fixing the feature generator weights after pretraining (\textit{Fixed Features}).
We pretrain for $350$ epochs for relevant methods and then train the full method for $150$ epoch. We use  Adam~\cite{kingma2014adam} with a learning rate of $1\text{e-}4$.
The hyperparameters were chosen to be the same as in the original Deep SVDD paper~\cite{ruff2018deep} for all experiments.

\subsection{Numerical Evaluation}\label{sec:eval}

\sisetup{
table-format = 2.2(1),
detect-weight
}

\begin{table*}
\caption{Different anomaly detection methods and sensor modality combinations evalauted using the \ac{AUROC}. We test in \sun{} conditions and indicate the standard deviation over 10 runs. \edited{The background color follows a gradient corresponding to the \ac{AUROC} \protect\scalerel*{\protect\includegraphics{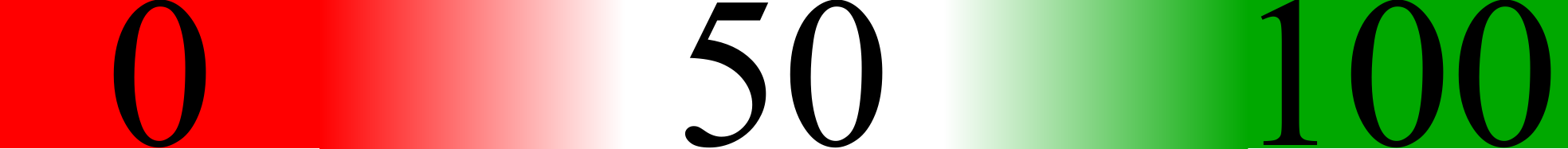}}{AUROC}. Addtionally, the highest performing combination is highlighted in bold text.} For a description of anomaly detection methods refer to \Cref{sec:anomaly_detection}. Modality short-hands are: \textit{RGB} - color, \textit{D} - Depth, \textit{G} - gravity aligned depth, \textit{N} - gravity aligned surface normals, \textit{A} - surface normal angle.}
\robustify\bfseries
\centering
\begin{tabular}{l|S S S S S S S S} 
 & \pbox{20cm}{\textbf{Autoencoder}} & \pbox{20cm}{\textbf{SVDD Soft}\\\textbf{No Pretraining}} & \pbox{20cm}{\textbf{SVDD Hard}\\\textbf{No Pretraining}} & \pbox{20cm}{\textbf{SVDD Soft}\\\textbf{Pretrained}} & \pbox{20cm}{\textbf{SVDD Hard}\\\textbf{Pretrained}} & \pbox{20cm}{\textbf{NVP No}\\\textbf{Pretraining}} & \pbox{20cm}{\textbf{NVP}\\\textbf{Pretrained}} & \pbox{20cm}{\textbf{NVP Fixed}\\\textbf{Features}}\\ \hline
RGB & 38.97+-0.39 \cellcolor[rgb]{1.00,0.89,0.89} & 18.55+-2.85 \cellcolor[rgb]{1.00,0.69,0.69} & 58.46+-12.89 \cellcolor[rgb]{0.83,0.94,0.83} & 60.09+-6.14 \cellcolor[rgb]{0.80,0.93,0.80} & 70.57+-3.95 \cellcolor[rgb]{0.59,0.86,0.59} & 73.52+-0.85 \cellcolor[rgb]{0.53,0.84,0.53} & 76.16+-2.95 \cellcolor[rgb]{0.48,0.83,0.48} & 64.34+-0.68 \cellcolor[rgb]{0.71,0.91,0.71} \\
D & 76.07+-0.30 \cellcolor[rgb]{0.48,0.83,0.48} & 80.04+-1.88 \cellcolor[rgb]{0.40,0.80,0.40} & 79.72+-2.38 \cellcolor[rgb]{0.41,0.80,0.41} & 48.89+-6.42 \cellcolor[rgb]{1.00,0.99,0.99} & 78.67+-2.02 \cellcolor[rgb]{0.43,0.81,0.43} & 77.69+-2.46 \cellcolor[rgb]{0.45,0.82,0.45} & 31.14+-0.93 \cellcolor[rgb]{1.00,0.81,0.81} & 81.41+-0.05 \cellcolor[rgb]{0.37,0.79,0.37} \\\hline
RGB+D & 63.92+-0.73 \cellcolor[rgb]{0.72,0.91,0.72} & 47.87+-7.03 \cellcolor[rgb]{1.00,0.98,0.98} & 50.40+-8.41 \cellcolor[rgb]{0.99,1.00,0.99} & 51.96+-17.67 \cellcolor[rgb]{0.96,0.99,0.96} & 71.79+-4.09 \cellcolor[rgb]{0.56,0.86,0.56} & 72.35+-3.02 \cellcolor[rgb]{0.55,0.85,0.55} & 74.84+-3.48 \cellcolor[rgb]{0.50,0.84,0.50} & 84.64+-0.21 \cellcolor[rgb]{0.31,0.77,0.31} \\
RGB+G & 68.44+-0.29 \cellcolor[rgb]{0.63,0.88,0.63} & 59.58+-5.39 \cellcolor[rgb]{0.81,0.94,0.81} & 70.77+-13.20 \cellcolor[rgb]{0.58,0.86,0.58} & 41.57+-27.75 \cellcolor[rgb]{1.00,0.92,0.92} & 74.64+-4.56 \cellcolor[rgb]{0.51,0.84,0.51} & 85.69+-1.43 \cellcolor[rgb]{0.29,0.76,0.29} & 83.12+-3.21 \cellcolor[rgb]{0.34,0.78,0.34} & 87.12+-0.85 \cellcolor[rgb]{0.26,0.76,0.26} \\
RGB+N & 92.92+-0.20 \cellcolor[rgb]{0.14,0.72,0.14} & 60.63+-30.35 \cellcolor[rgb]{0.79,0.93,0.79} & 43.39+-6.22 \cellcolor[rgb]{1.00,0.93,0.93} & 40.00+-3.41 \cellcolor[rgb]{1.00,0.90,0.90} & 53.44+-9.55 \cellcolor[rgb]{0.93,0.98,0.93} & 86.45+-0.86 \cellcolor[rgb]{0.27,0.76,0.27} & 45.04+-13.28 \cellcolor[rgb]{1.00,0.95,0.95} & 93.12+-1.00 \cellcolor[rgb]{0.14,0.72,0.14} \\
RGB+A & 67.79+-0.08 \cellcolor[rgb]{0.64,0.88,0.64} & 20.27+-7.26 \cellcolor[rgb]{1.00,0.70,0.70} & 63.17+-6.89 \cellcolor[rgb]{0.74,0.91,0.74} & 36.09+-33.44 \cellcolor[rgb]{1.00,0.86,0.86} & 69.44+-4.81 \cellcolor[rgb]{0.61,0.87,0.61} & 90.06+-1.10 \cellcolor[rgb]{0.20,0.74,0.20} & 68.43+-12.04 \cellcolor[rgb]{0.63,0.88,0.63} & 87.69+-0.14 \cellcolor[rgb]{0.25,0.75,0.25} \\\hline
D+N & 92.81+-0.09 \cellcolor[rgb]{0.14,0.72,0.14} & 75.40+-24.53 \cellcolor[rgb]{0.49,0.83,0.49} & 57.76+-7.20 \cellcolor[rgb]{0.84,0.95,0.84} & 89.30+-1.50 \cellcolor[rgb]{0.21,0.74,0.21} & 62.27+-10.90 \cellcolor[rgb]{0.75,0.92,0.75} & 83.93+-1.18 \cellcolor[rgb]{0.32,0.78,0.32} & 49.72+-10.93 \cellcolor[rgb]{1.00,1.00,1.00} & 90.08+-0.97 \cellcolor[rgb]{0.20,0.74,0.20} \\
D+A & 79.91+-0.11 \cellcolor[rgb]{0.40,0.80,0.40} & 80.63+-2.55 \cellcolor[rgb]{0.39,0.80,0.39} & 61.91+-0.56 \cellcolor[rgb]{0.76,0.92,0.76} & 49.43+-1.47 \cellcolor[rgb]{1.00,0.99,0.99} & 70.10+-3.59 \cellcolor[rgb]{0.60,0.87,0.60} & 77.44+-0.71 \cellcolor[rgb]{0.45,0.82,0.45} & 55.69+-2.78 \cellcolor[rgb]{0.89,0.96,0.89} & 84.12+-0.57 \cellcolor[rgb]{0.32,0.77,0.32} \\
G+A & 80.72+-0.17 \cellcolor[rgb]{0.39,0.80,0.39} & 56.94+-21.08 \cellcolor[rgb]{0.86,0.95,0.86} & 71.07+-7.44 \cellcolor[rgb]{0.58,0.86,0.58} & 52.79+-20.09 \cellcolor[rgb]{0.94,0.98,0.94} & 78.53+-7.80 \cellcolor[rgb]{0.43,0.81,0.43} & 87.47+-0.42 \cellcolor[rgb]{0.25,0.75,0.25} & 80.14+-1.45 \cellcolor[rgb]{0.40,0.80,0.40} & 86.29+-0.84 \cellcolor[rgb]{0.27,0.76,0.27} \\\hline
RGB+D+N & 94.17+-0.27 \cellcolor[rgb]{0.12,0.71,0.12} & 44.28+-21.54 \cellcolor[rgb]{1.00,0.94,0.94} & 51.99+-2.13 \cellcolor[rgb]{0.96,0.99,0.96} & 52.48+-5.61 \cellcolor[rgb]{0.95,0.98,0.95} & 54.28+-5.62 \cellcolor[rgb]{0.91,0.97,0.91} & 85.50+-2.28 \cellcolor[rgb]{0.29,0.77,0.29} & 46.29+-14.02 \cellcolor[rgb]{1.00,0.96,0.96} & 94.99+-0.41 \cellcolor[rgb]{0.10,0.70,0.10} \\
RGB+D+A & 76.19+-0.44 \cellcolor[rgb]{0.48,0.83,0.48} & 44.13+-12.67 \cellcolor[rgb]{1.00,0.94,0.94} & 55.39+-14.01 \cellcolor[rgb]{0.89,0.96,0.89} & 41.16+-20.11 \cellcolor[rgb]{1.00,0.91,0.91} & 69.45+-5.57 \cellcolor[rgb]{0.61,0.87,0.61} & 89.45+-1.23 \cellcolor[rgb]{0.21,0.74,0.21} & 82.16+-3.77 \cellcolor[rgb]{0.36,0.79,0.36} & 90.11+-0.71 \cellcolor[rgb]{0.20,0.74,0.20} \\
RGB+G+N & 94.51+-0.04 \cellcolor[rgb]{0.11,0.71,0.11} & 38.95+-35.13 \cellcolor[rgb]{1.00,0.89,0.89} & 61.76+-4.43 \cellcolor[rgb]{0.76,0.92,0.76} & 38.35+-10.48 \cellcolor[rgb]{1.00,0.88,0.88} & 62.28+-3.63 \cellcolor[rgb]{0.75,0.92,0.75} & 87.53+-2.10 \cellcolor[rgb]{0.25,0.75,0.25} & 50.42+-11.81 \cellcolor[rgb]{0.99,1.00,0.99} & \bfseries 95.14+-1.47 \cellcolor[rgb]{0.10,0.70,0.10} \\
RGB+G+A & 78.77+-0.54 \cellcolor[rgb]{0.42,0.81,0.42} & 58.44+-10.18 \cellcolor[rgb]{0.83,0.94,0.83} & 73.67+-9.14 \cellcolor[rgb]{0.53,0.84,0.53} & 33.76+-28.66 \cellcolor[rgb]{1.00,0.84,0.84} & 75.25+-5.44 \cellcolor[rgb]{0.50,0.83,0.50} & 92.85+-0.26 \cellcolor[rgb]{0.14,0.72,0.14} & 80.06+-2.36 \cellcolor[rgb]{0.40,0.80,0.40} & 91.60+-0.16 \cellcolor[rgb]{0.17,0.73,0.17} \\

\end{tabular}
\label{tab:auroc}
\end{table*}

For quantitative analysis of anomaly detection approaches and sensor modality combinations, we train our approaches on the base training set and use data from one \sun{} sortie as test set.
We use the threshold-independent \ac{AUROC} as performance metric.
\Cref{tab:auroc} shows results for all evaluated approaches and sensor modality combinations. 

\subsubsection{Anomaly Detection Methods}

\begin{figure}[t]
  \begin{subfigure}[t]{0.49\linewidth}
    \centering
    \includegraphics[width=\linewidth]{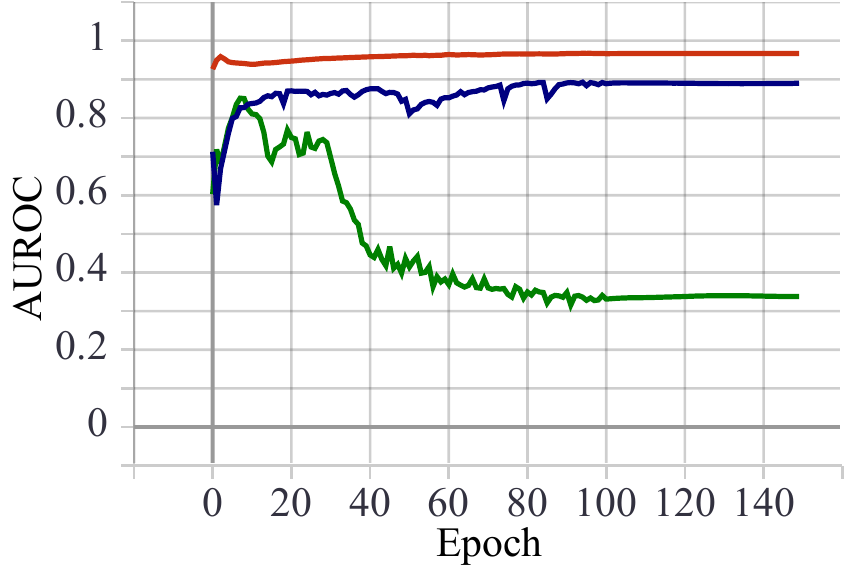}
    \caption{Test Set \ac{AUROC}}
  \end{subfigure}
  \hfill
  \begin{subfigure}[t]{0.49\linewidth}
    \centering
    \includegraphics[width=\linewidth]{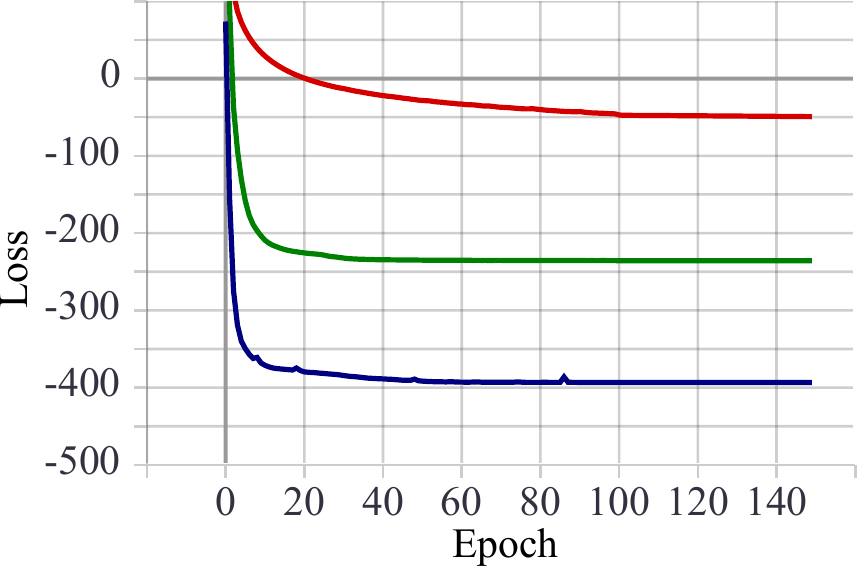}
    \caption{Training Loss}
  \end{subfigure}
  \caption{Training curves of NVP methods show that high likelihood of the training distribution does not correlate to high discriminative performance on the test set. \textit{Blue} - No Pretraining, \textit{Green} - Pretrained, \textit{Red} - Pretrained + fixed feature generator}
  \label{fig:auc_loss}
\end{figure}

We can see that the Real-NVP based approaches clearly outperform the autoencoder and Deep SVDD approaches. 
This approach tries to explicitly learn the posterior distribution of traversable image features and allows us to learn arbitrary distributions, whereas Deep SVDD assumes a uni-modal distribution, since it classifies all features inside a hypersphere as inliers. 
Additionally, the objective function does not force the network to learn the joint distribution over all input modalities. 
It can in principle converge to a solution which ignores some input modalities if others allow easier mapping to a fixed feature point.
\edited{The autoencoder approach is able to learn a good approximation of the underlying distribution, as evidenced by generally higher performance than Deep SVDD which rivals Real-NVP, when provided with surface normals. 
Its otherwise inferior performance to Real-NVP stems from the appearance-based decision criterion, which is a poor similarity measure.}
The higher performance of the Real-NVP version trained with fixed feature generator weights we assume to be caused by joint distribution learning of multiple modalities.
Some parts of the actual underlying distribution are ignored without fixed features, in favor of mapping to a simpler posterior distribution, where higher likelihood can be achieved. 
An indicator is a lower training loss while also having a lower discriminative performance pictured in \Cref{fig:auc_loss}. 

\subsubsection{Sensor Modalities}

Unsurprisingly, geometric modalities enable consistently high performance, given that it is the preferred modality for traversability classification in literature~\cite{wermelinger2016navigation, krusi2017driving}.
Providing explicit surface normal information (\textit{N}) provides significant gains over depth (\textit{D}) and gravity-aligned depth (\textit{G}) hinting that the convolutional layers of the feature generator do not learn to fully leverage the presented geometric information. 
Interestingly, the surface normal angle (\textit{A}), which directly corresponds to terrain inclination, commonly used for traversability estimation in analytical approaches~\cite{wermelinger2016navigation}, does not provide the same performance boost as the normal vectors. 

Using \textit{RGB}-only shows significantly worse performance than any combination with geometric information. 
This is not surprising, given that geometry is a major deciding factor in whether or not terrain is traversable.
Inferring geometric information from color images is a hard problem even when networks are explicitly trained for this task which makes it unlikely that our network learns to reason about it.
Hence, the network cannot distinguish between concrete walls and asphalt streets, which have very similar texture and are common in our dataset.

However, in many cases geometry alone is not enough to infer traversability. 
For example, tall grass leads to a geometry that suggests untraversable, but can easily be recognized as traversable from the RGB image. 
Adding \textit{RGB} information to any geometric modality combination improves performance, because it helps to distinguish rigid from non-rigid geometry and gives additional information in image regions with missing depth due to stereo matching failure. 

\begin{figure}[t]
  \begin{subfigure}[t]{0.49\linewidth}
    \centering
    \includegraphics[width=\linewidth]{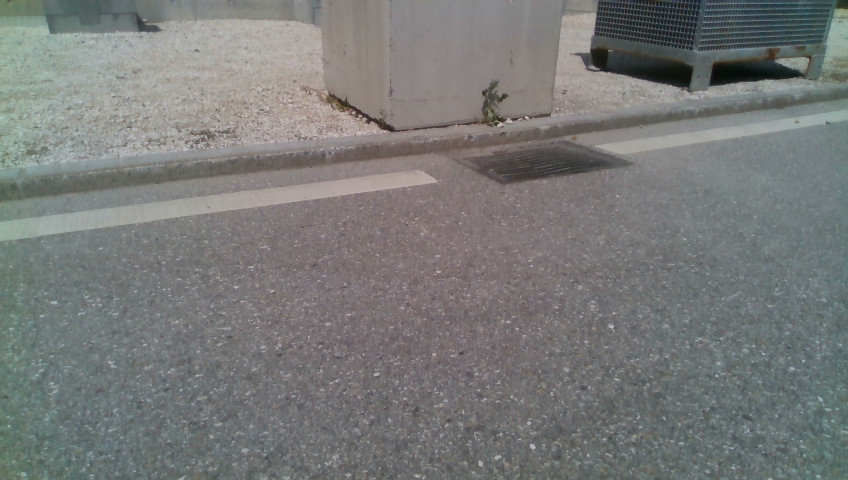}
    \caption{RGB - \sun{} concrete.}
  \end{subfigure}
  \hfill
  \begin{subfigure}[t]{0.49\linewidth}
    \centering
    \includegraphics[width=\linewidth]{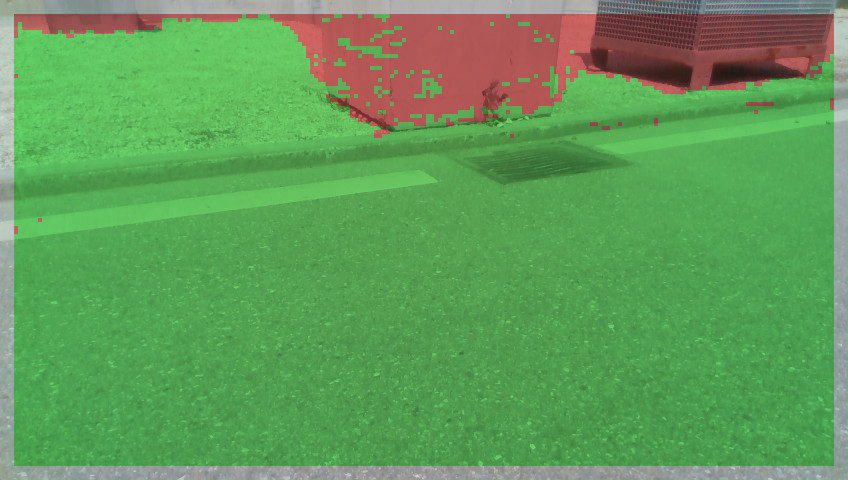}
    \caption{Anomaly Mask - \sun{} concrete.}
  \end{subfigure}
  \begin{subfigure}[t]{0.49\linewidth}
    \centering
    \includegraphics[width=\linewidth]{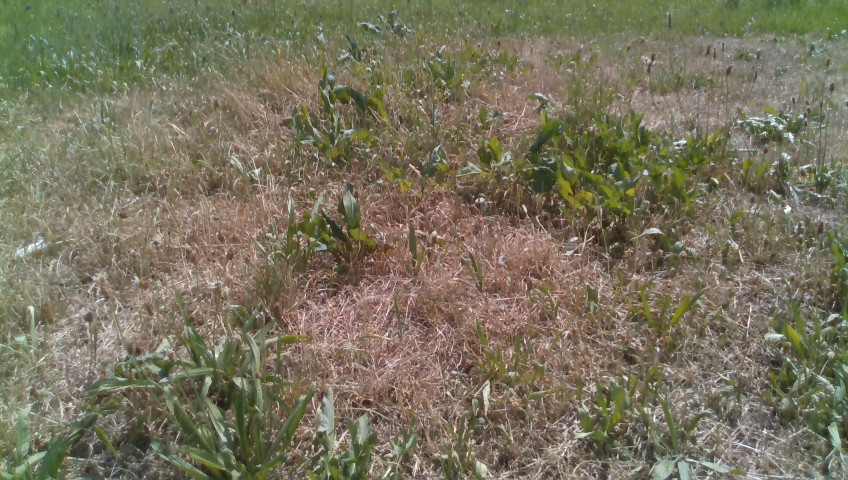}
    \caption{RGB - \sun{} grass.}
  \end{subfigure}
  \hfill
  \begin{subfigure}[t]{0.49\linewidth}
    \centering
    \includegraphics[width=\linewidth]{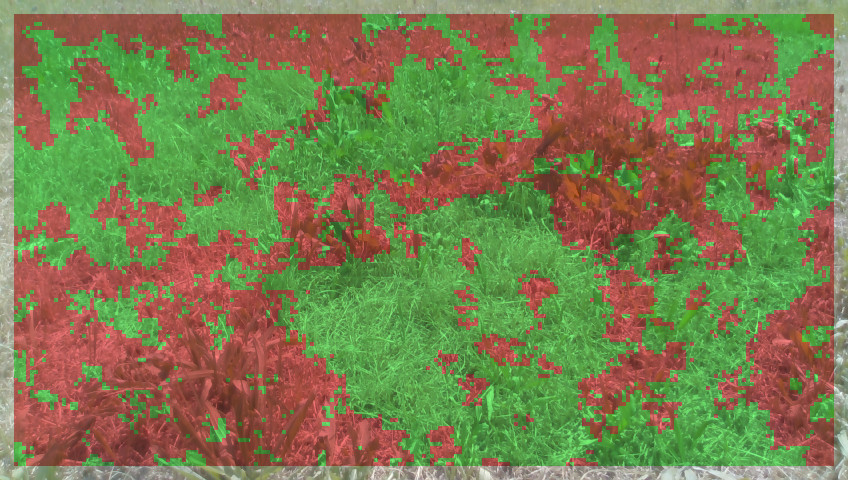}
    \caption{Anomaly Mask - \sun{} grass.}
  \end{subfigure}
  \begin{subfigure}[t]{0.49\linewidth}
    \centering
    \includegraphics[width=\linewidth]{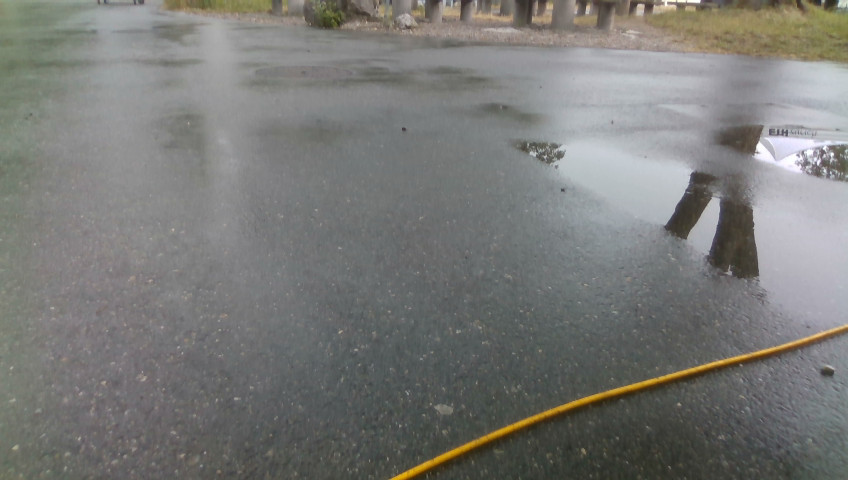}
    \caption{RGB - \rain{} reflections.}
  \end{subfigure}
  \hfill
  \begin{subfigure}[t]{0.49\linewidth}
    \centering
    \includegraphics[width=\linewidth]{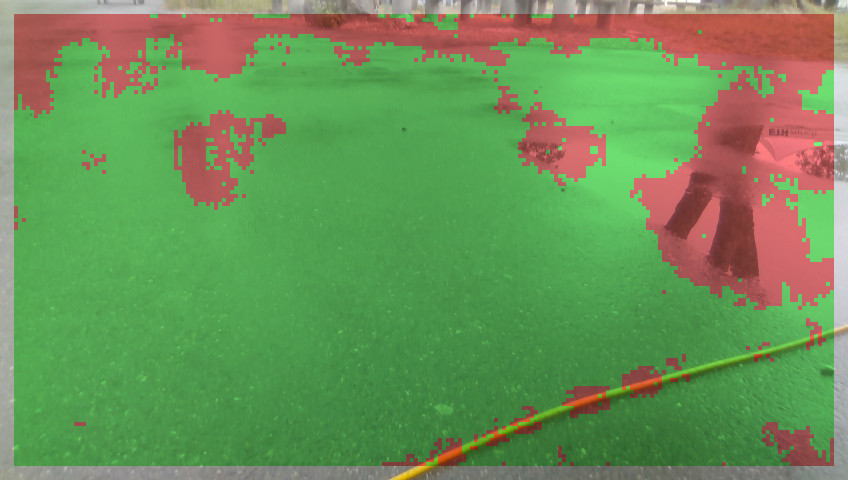}
    \caption{Anomaly Mask - \rain{} reflections.}
  \end{subfigure}
  \begin{subfigure}[t]{0.49\linewidth}
    \centering
    \includegraphics[width=\linewidth]{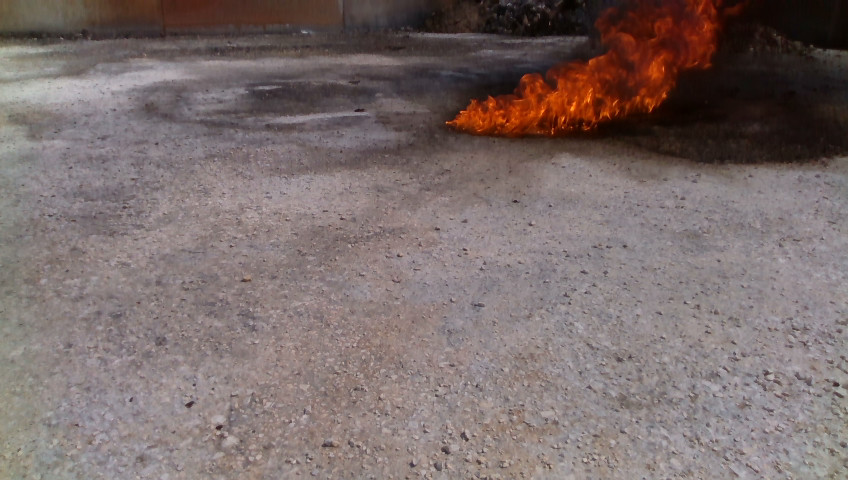}
    \caption{RGB - \fire{}.}
  \end{subfigure}
  \hfill
  \begin{subfigure}[t]{0.49\linewidth}
    \centering
    \includegraphics[width=\linewidth]{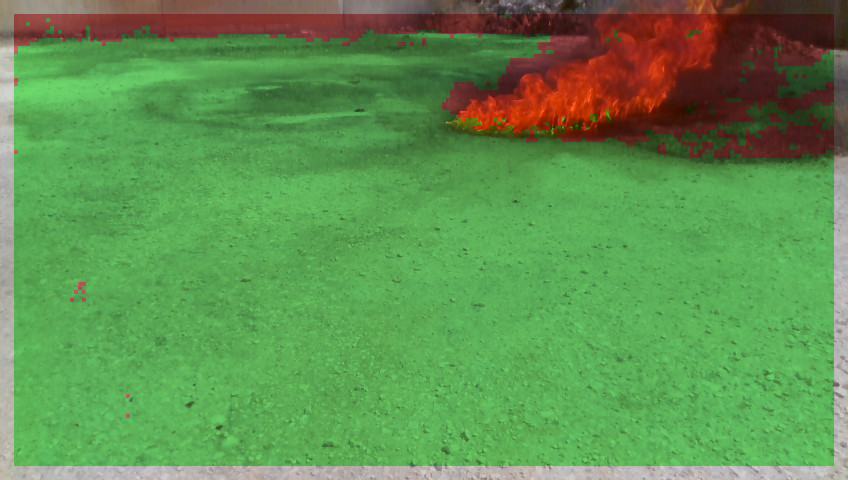}
    \caption{Anomaly Mask - \fire{}.}
  \end{subfigure}
  \caption{Qualitative results of anomaly detection in different environments. (a)+(b) Depth information helps to classify an ambiguous texture as obstacle. (c)+(d) While short grass is present in the training set, taller vegetation is not and gets classified as outlier. (e)+(f) Specular reflections which are not in the training set are classified as outliers. (g)+(f) Fire as an unknown obstacle is clearly identified.}
  \label{fig:qualitative}
\end{figure}

Qualitative results of the highest performing method, \textit{Real-NVP Fixed Features} with \textit{RGB+G+N} are shown in \Cref{fig:qualitative}.

\subsection{Incremental Learning}

In this section we will demonstrate how adding more input data from new environments allows our method to scale and improve performance over time.
We use the modality combination without any post-processing of the depth information, \textit{RGB+D}, and Real-NVP on the latent space of a feature generator with fixed weights that were pretrained with an autoencoder.
We will train our network with increasingly more data and see how the performance evolves for different environmental conditions. 
We reduce the training data set size from $~10\,000$ to $~500$ image patches, which will serve as base training set, while we incrementally add $~500$ image patches of one sortie in a given condition to the training data.
The second sortie under that condition will serve as the test set.
We use the shorter sortie under \rain{} conditions as train set and compensate for the shorter duration by sampling multiple patches per image to reach $500$ samples. 
Additionally, \wet{} and \fire{} conditions will remain purely test sets as we only have data of a single sortie.
Because of the reduced dataset size we only train for $10$ epochs. 

\begin{figure*}[t]
\centering
  \begin{subfigure}[t]{0.325\linewidth}
    \centering
    \includegraphics[width=\linewidth]{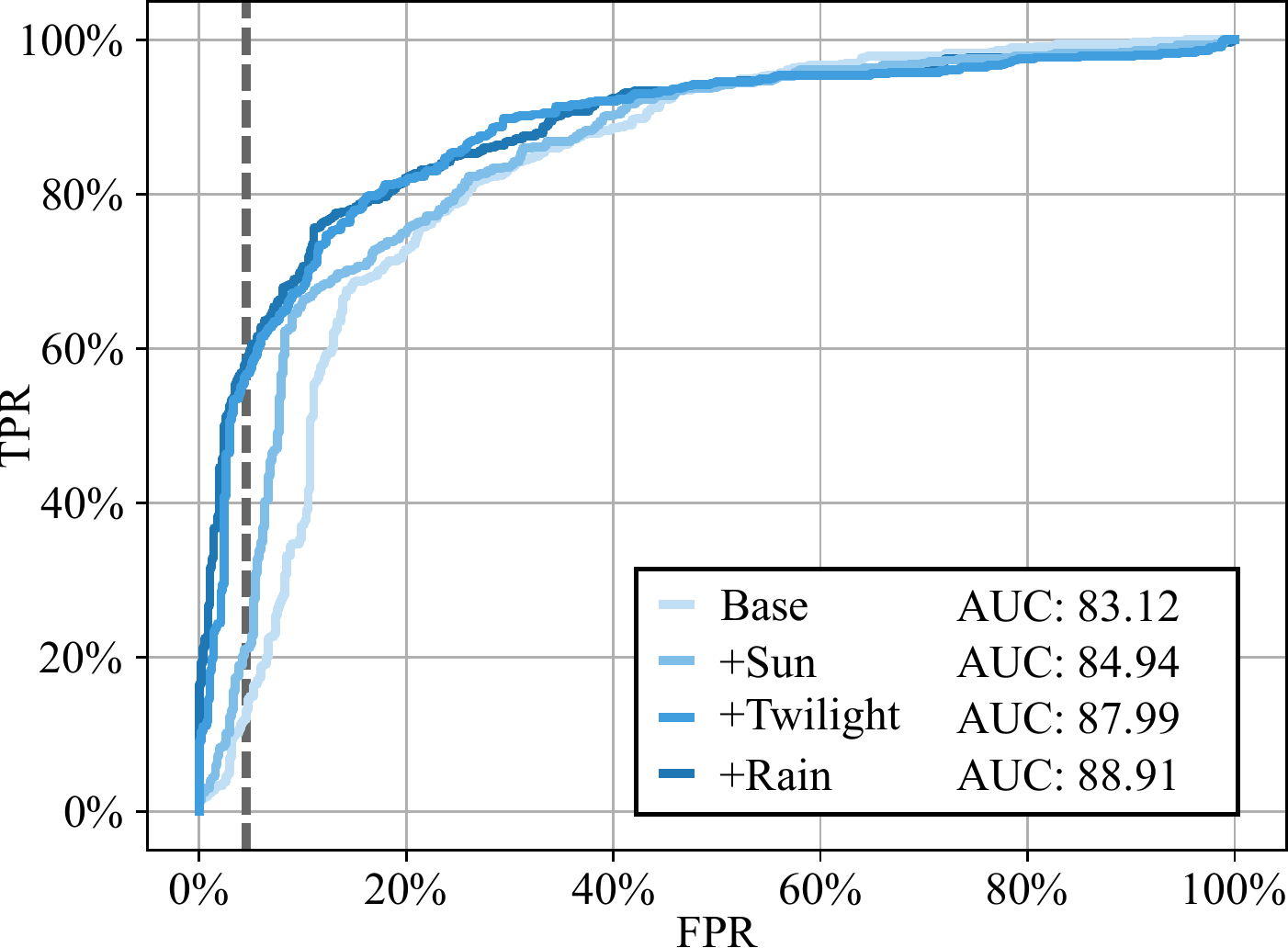}
    \caption{\sun{}}
  \end{subfigure}
  \hfill
  \begin{subfigure}[t]{0.325\linewidth}
    \centering
    \includegraphics[width=\linewidth]{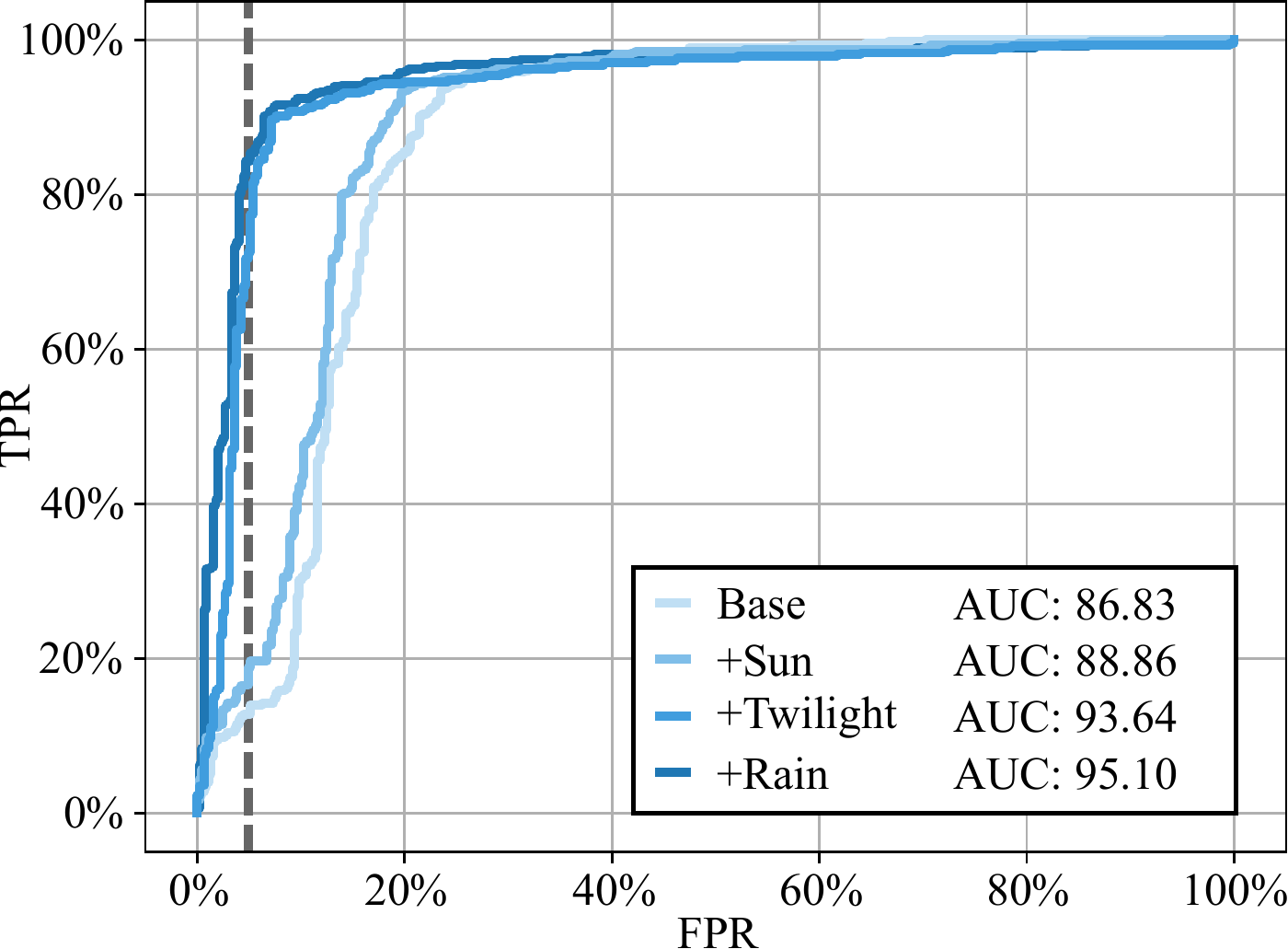}
    \caption{\twilight{}}
  \end{subfigure}  
  \begin{subfigure}[t]{0.325\linewidth}
    \centering
    \includegraphics[width=\linewidth]{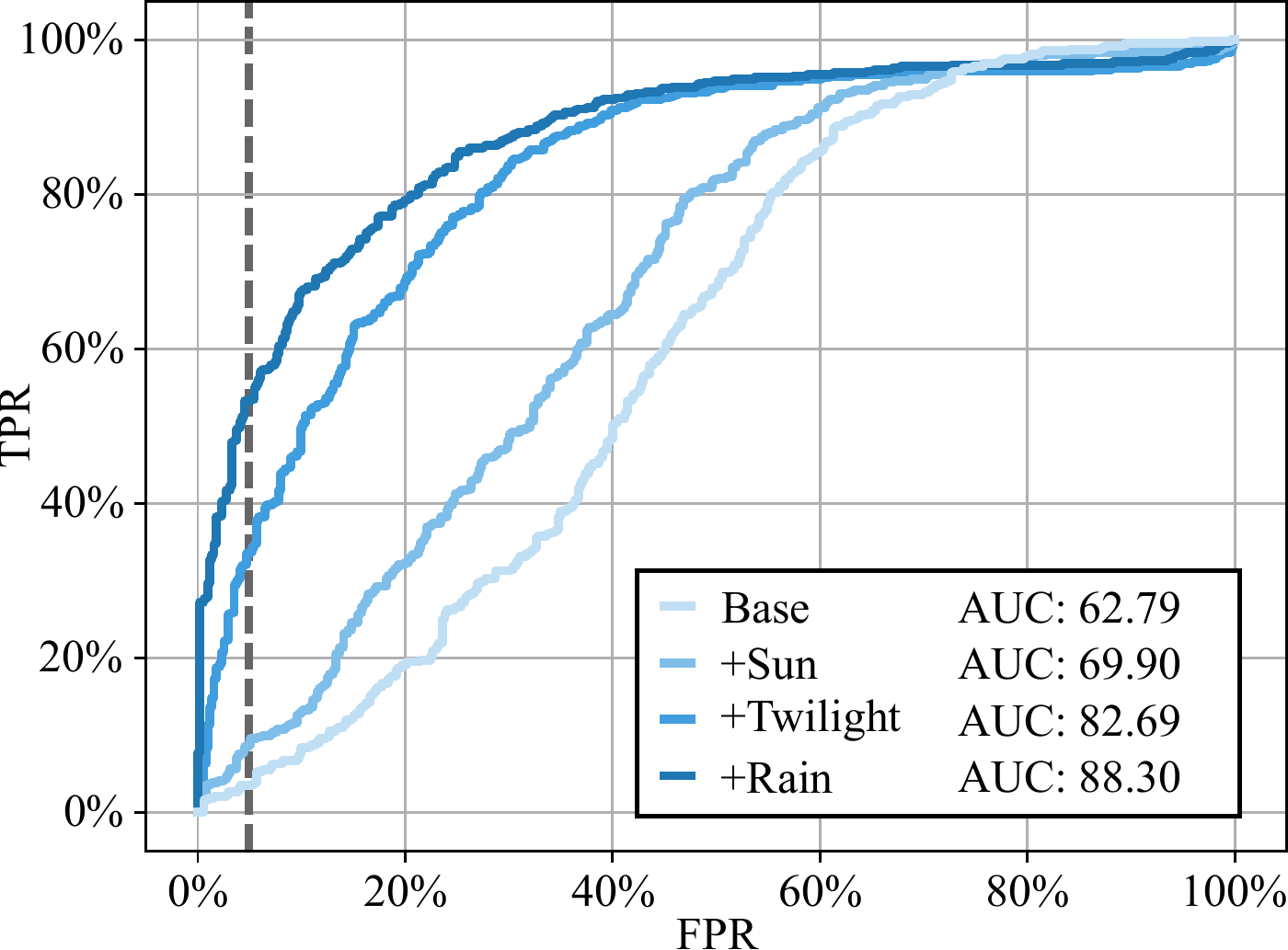}
    \caption{\rain{}}
  \end{subfigure}
  \begin{subfigure}[t]{0.325\linewidth}
    \centering
    \includegraphics[width=\linewidth]{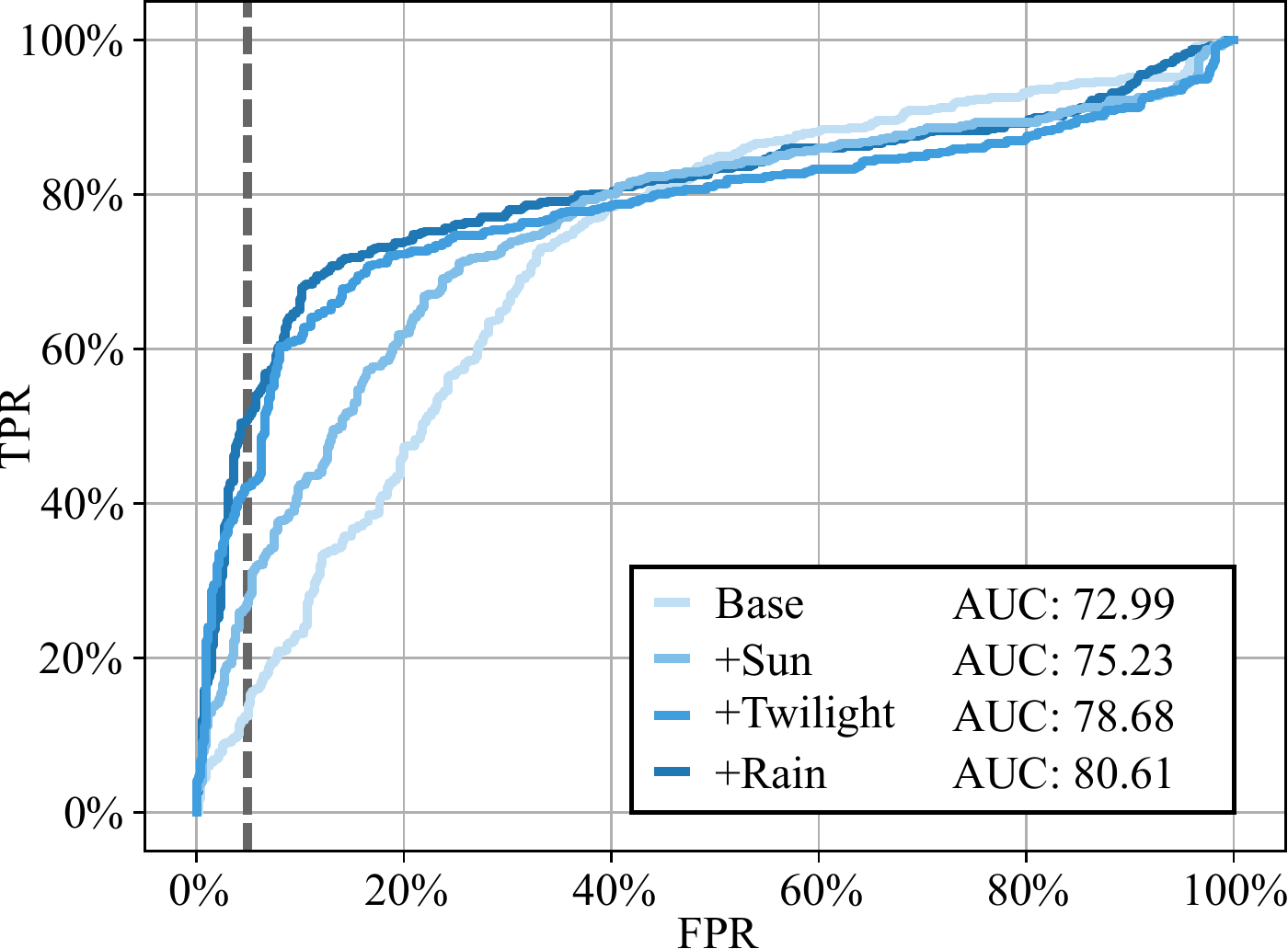}
    \caption{\wet{}}
  \end{subfigure}
  \begin{subfigure}[t]{0.325\linewidth}
    \centering
    \includegraphics[width=\linewidth]{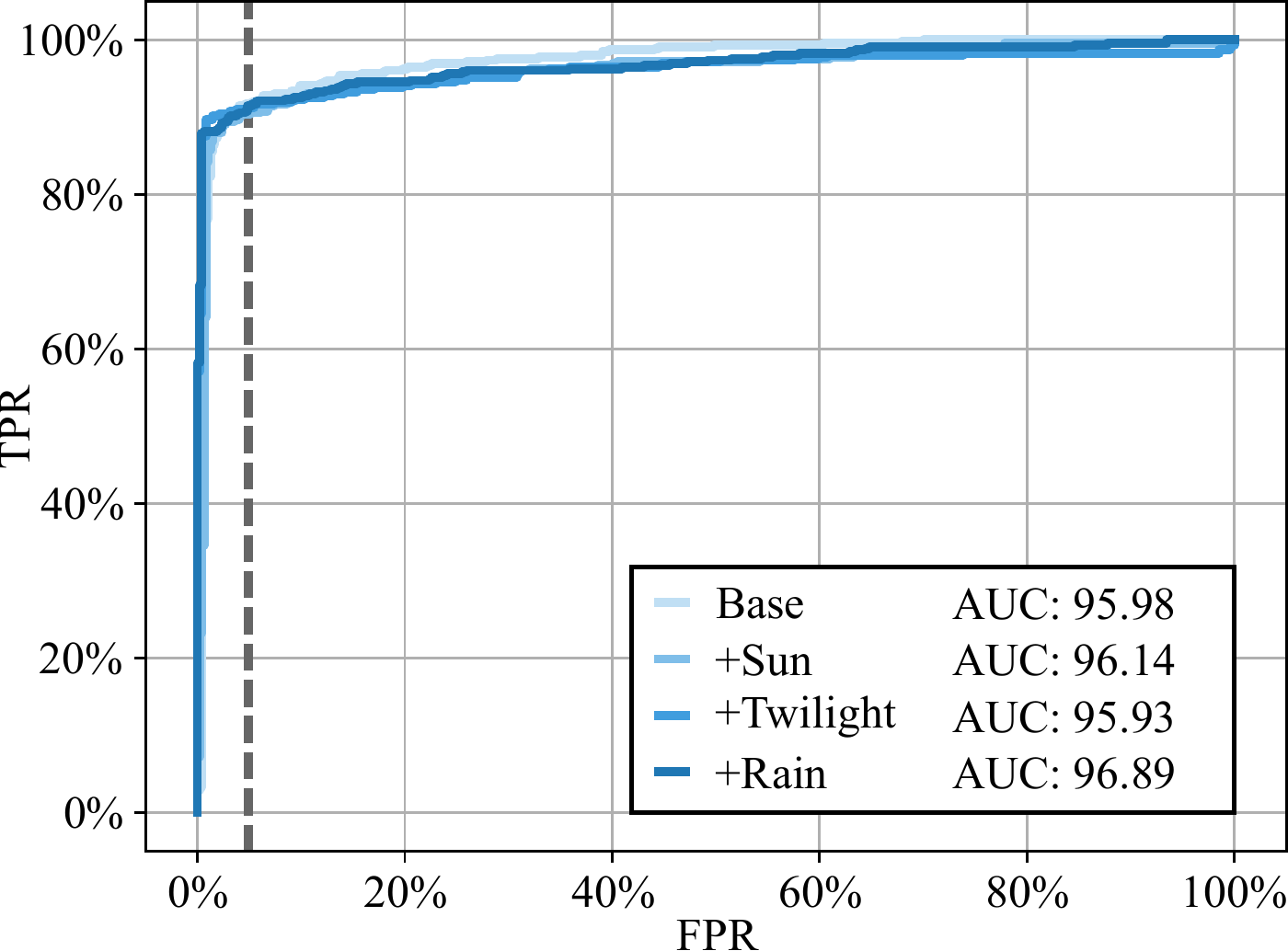}
    \caption{\fire{}}
  \end{subfigure}
  \caption{ROC curves for \textit{NVP Fixed Features RGB+D} on test sets trained with incrementally more data. The $5$\%FPR threshold is indicates by a dashed grey line. The curves shift towards the left with more data, which implies improved performance at low false-positive rates. Note that a low false-positive rate is our desired operating domain as false positives can cause catastrophic failure.}
  \label{fig:roc_incremental}
\end{figure*}

The results in \Cref{fig:roc_incremental} show an increase in performance for all conditions when adding more data. 
With additional data, the \ac{AUROC} for \sun{} conditions improves over the results with the full training set shown in \Cref{tab:auroc}, which was tested with the same data.
The highest gains are achieved for \rain{} conditions, which shows that diverse training data is crucial to handle operation in various environmental conditions.
Strong performance on \fire{}, where unsafe terrain is dominated by bright fire and billows of black smoke, show that our anomaly detection based approach can safely handle unknown environmental hazards.
An important additional note is, that the true-positive rate at $5$\% false-positive rate (\tpr{}) improves drastically for all but \fire{} conditions, when adding more data. 
It also improves over the full training data \tpr{} of $43$\% for \sun{} conditions.
This measure is a good indicator for a navigation task operating point, since we want a low false-positive rate to minimize the chance of catastrophic robot failure.
Interestingly, \rain{} and \twilight{} data seem to be much more significant for improving \tpr{} than \sun{} data, even when evaluating under \sun{} conditions.

\edited{
\subsection{Network Inference Time}

All networks run in real-time on mobile computation hardware.
\Cref{tab:inference} reports inference times once per base approach, since the different training methods do not alter the inference time.
}

\sisetup{
table-format=3.1
}

\begin{table}[h!]
\caption{Inference times for the three base approaches on an Nvidia Jetson Xavier (15W mode) with input image size $848\times480$.}
\centering
\begin{tabular}{r|S S S} 
RGB+G+N & \textbf{Autoencoder} & \textbf{SVDD} & \textbf{Real-NVP} \\ \hline
time [\si{ms}] & 9.4 & 4.6 & 42.9 \\
rate [\si{\hertz}] & 106.1 & 216.6 & 23.3 \\
\end{tabular}
\label{tab:inference}
\end{table}

%===============================================================================

\section{Conclusion}
\label{sec:conclusion}

In this work we demonstrated a method for safe robot navigation in the presence of unknown obstacles using anomaly detection.
Our approach combining a feature embedding with normalizing flow is able to operate in a variety of environments and scales well with additional data. Our semi-supervised data collection pipeline enables to collect multi-modal data from experience without any manual labelling.
The highest performance was achieved with a sensor modality combination of RGB images, depth and surface normals.
Our work opens up several avenues for future research.
An active exploration approach could automate the collection of new data and ease the expansion of the robot's operating range.
While the current approach trains only on samples of safe terrain, extending it to use sparse experiences of robot failure could sharpen decision boundaries in ambiguous environments.
Additionally, increasing the receptive field size of our network could allow the robot to reason about even more complex environments where translucent and reflective objects are present.

%===============================================================================

% \clearpage

%===============================================================================

\bibliographystyle{IEEEtran}
\bibliography{IEEEabrv,references}

\end{document}